\documentclass[10pt,twocolumn]{article}
\usepackage{times}
\usepackage{graphicx}
\usepackage{amsbsy}
\usepackage{amsfonts}
\usepackage{rotating}
\usepackage{booktabs}
\usepackage{listings}
\usepackage{algorithmic,algorithm}
\usepackage{url}
\usepackage{verbatim}
\begin{document}

\title{Local Optima Networks\\ of the Quadratic Assignment Problem
\thanks{Fabio Daolio and Marco Tomassini are with the Information Systems Department,
Faculty of Business and Economics, University of Lausanne, Lausanne, Switzerland. %Emails: \texttt{\{fabio.daolio,marco.tomassini\}@unil.ch}.
S\'ebastien Verel is with INRIA Lille - Nord Europe and University of Nice Sophia-Antipolis~/~CNRS, Nice, France.
Gabriela Ochoa is with the Automated Scheduling, Optimisation and Planning (ASAP) Group, School of Computer Science, University of Nottingham, Nottingham, UK.
}
}

\author{Fabio Daolio, S\'ebastien Verel, Gabriela Ochoa, Marco Tomassini}

\maketitle
\thispagestyle{empty}

\begin{abstract}

Using a recently proposed model for combinatorial landscapes, {\em Local Optima Networks (LON)}, we conduct a thorough analysis of two types of instances of the Quadratic Assignment Problem (QAP). This network model is a reduction of the landscape in which the nodes correspond to the local optima, and the edges account for the notion of adjacency between their basins of attraction. The model was inspired by the notion of `inherent network' of potential energy surfaces proposed in physical-chemistry. The local optima networks extracted from the so called {\em uniform} and {\em real-like} QAP instances, show features clearly distinguishing these two types of instances. Apart from a clear confirmation that the search difficulty increases with the problem dimension, the analysis provides new confirming evidence explaining  why the real-like instances are easier to solve exactly using heuristic search, while the uniform instances are easier to solve approximately.  Although the local optima network model is still under development, we argue that it provides a novel view of combinatorial landscapes, opening up the possibilities for new analytical tools and understanding of problem difficulty in combinatorial optimization.
\end{abstract}

\section{Introduction}
\label{sec:intro}

In a series of papers we introduced a novel representation for combinatorial landscapes that we called \textit{local optima network}~\cite{gecco08,alife08,pre09}. This is a view of landscapes derived from a previously proposed one for continuous energy landscapes by Doye~\cite{doye02,doye05} but it has been modified and adapted to work for discrete combinatorial spaces. It is based on the idea of compressing the information given by the whole problem configuration space into a smaller mathematical object which is the graph having as vertices the optima configurations of the problem and as edges the possible weighted transitions between these optima. The methodology is intended to be a descriptive one in the first place; for example, some measures on the optima networks have been found to be related with problem difficulty. In the longer term, the methodology is expected to be useful for suggesting improvements in local search heuristics and perhaps even for suggesting new ones. In recent work we have studied by this method the family of Kauffman's $NK$-landscapes~\cite{kauffman93}, for which we have shown that some optima network statistics can be related to the tunable difficulty of these landscapes. Moreover, since to obtain the local optima of the configuration space we need to explore the corresponding basins, the above graph is also a description of the basins and of their connectivity. In this way we have been able to find previously
known properties of these basins, as well as new ones~\cite{alife08,pre09}. However, although useful for classification purposes, the $NK$ family of landscapes is a highly artificial one. For this reason, in the present paper we study the more realistic problem called the \textit{quadratic assignment problem} (QAP). The quadratic assignment problem, as introduced by Koopmans and Beckmann~\cite{1957} in 1957, is a combinatorial optimization problem which is known to be NP-hard~\cite{sahni1976p}. This paper presents preliminary results of an exhaustive analysis of small instances' fitness landscapes by means of extracting the networks of local optima and evaluating their statistics.

The QAP deals with the relative location of units that interact with one another. The objective is to minimize the total cost of interactions. The problem can be stated in this way: there are $n$ units or facilities to be assigned to $n$ predefined locations, where each location can accommodate any one unit; location $i$ and location $j$ are separated by a distance $a_{ij}$, generically representing the per unit cost of interaction between the two locations; a flow of value $b_{ij}$ has to go from unit $i$ to unit $j$; the objective is to find and assignment, i.e. a bijection from the set of facilities onto the set of locations, which minimizes the sum of products flow $\times$ distance.
Mathematically it can be formulated as:
\begin{equation}
\min_{\pi \in P(n)} C(\pi)=\sum_{i=1}^{n}\sum_{j=1}^{n}{a_{ij}b_{\pi_{i}\pi_{j}}}
\label{eq:fitness}
\end{equation}

\noindent where $A=\{a_{ij}\}$ and $B=\{b_{ij}\}$ are the two $n \times n$ distance and flow matrixes, $\pi_{i}$ gives the location of facility $i$ in permutation $\pi \in P(n)$, and $P(n)$ is the set of all permutations of $\{1,2,...,n\}$, i.e. the QAP search space.
The structure of the distance and flow matrices characterize the class of instances of the QAP problem. Later in the article it is explained which are the classes of instances used in the present work.

The paper is structured as follows. The next section gives a number of concepts, definitions and
algorithms  used to obtain and describe the optima networks of the QAP problem. Section \ref{sec:analysis} discusses the analysis of the network data thus obtained. The discussion applies to both the optima graph, as well as to the associated basins of attraction. Finally,  section \ref{sec:discs} presents our conclusions and suggestions for further work.

\section{Definitions and Algorithms}
\label{sec:defs}

Given a fitness landscape for an instance of the QAP problem, we have to define the associated optima network by providing definitions for the nodes and the edges of the network. The vertexes of the graph can be straightforwardly defined as the local minima of the landscape. This  work is a first step toward the network characterization of QAP landscapes: we present the analysis of  small instances (see Sect.~\ref{sec:experimentalSetting}). For these instances, it is feasible to obtain the nodes of the graph exhaustively by running a best-improvement local search algorithm from every configuration of the search space as described below. Before explaining how the edges of the network are obtained,  a number of relevant definitions are summarized.

A Fitness landscape~\cite{stadler-02} is a triplet $(S, V, f)$ where $S$ is a set of potential solutions i.e. a search space, $V : S \longrightarrow 2^S$, a neighborhood structure, is a function that assigns to every $s \in S$ a set of neighbors $V(s)$, and $f : S \longrightarrow \mathbb{R}$ is a fitness function that can be pictured as the \textit{height} of the
corresponding solutions. For the QAP problem, a search space configuration  is a permutation of the facility locations
of length $n$, therefore the search space size is $n!$.  The neighborhood of a configuration is defined
by the pairwise exchange operation, which is the most basic operation used by many meta-heuristics for QAP. This operator simply exchanges any two positions in a permutation, thus transforming it  into another permutation. The neighborhood size is thus $|V(s)| = n(n-1)/2$. 
Finally the fitness value of a solution can be simply set to be equal to the opposite of the assignment cost defined in eq.~\ref{eq:fitness}.

The $HillClimbing$ algorithm used to determine the local optima and
therefore define the basins of attraction, is given in Algorithm~\ref{algoHC}. It defines a mapping from the
search space $S$ to the set of locally optimal solutions $S^*$, where a local optimum is a solution $s^{*}$ such that $\forall  s \in
V(s^{*})$, $f(s) < f(s^{*})$.

\begin{algorithm}
\caption{{\em Hill-Climbing}} \label{algoHC}
\begin{algorithmic}
\STATE Choose initial solution $s \in \cal S$ \REPEAT
    \STATE choose $s^{'} \in V(s)$ such that $f(s^{'}) = max_{x \in V(s)}\ f(x)$
        \IF{$f(s) < f(s^{'})$}
            \STATE $s \leftarrow s^{'}$
    \ENDIF
\UNTIL{$s$ is a  Local optimum}
\end{algorithmic}
\end{algorithm}

The basin of attraction of a local optimum $i \in S$ is the set $b_i = \{
s \in S ~|~ HillClimbing(s) = i \}$. The size of the basin of
attraction of a local optimum $i$ is the cardinality of $b_i$.
Notice that for non-neutral fitness landscapes, the basins of attraction as defined above produce a
partition of the configuration space $S$. Therefore, $S = \cup_{i
\in S^{*}} b_i$ and $\forall i \in S$ $\forall j \not= i$, $b_i \cap
b_j = \emptyset$.\\
We can now define the edge of a weight that connects two feasible solutions in the
 fitness landscape.
\noindent For each pair of solutions $s$ and $s^{'}$, $p(s
\rightarrow s^{'} )$ is the probability
to pass from $s$ to $s^{'}$ with the given neighborhood structure.
For the search space of permutations of $n$ elements, and the pairwise exchange operation, there are $n(n-1)/2$ neighbors for each solution, therefore:

\noindent if $s^{'} \in V(s)$ , $p(s \rightarrow s^{'} ) = \frac{1}{n(n-1)/2}$ and \\
if $s^{'} \not\in V(s)$ , $p(s \rightarrow s^{'} ) = 0$.

\noindent The probability to pass from a solution $s
\in S$ to a solution belonging to the basin $b_j$, is defined as:
$$
p(s \rightarrow b_j ) = \sum_{s^{'} \in b_j} p(s \rightarrow s^{'} )
$$

\noindent Notice that $p(s \rightarrow b_j ) \leq 1$.
Thus, the total probability of going from basin $b_i$ to
basin $b_j$ is the average over all $s \in b_i$ of the transition
probabilities  to solutions $s^{'} \in b_j$ :

$$p(b_i \rightarrow b_j) = \frac{1}{\sharp b_i} \sum_{s \in b_i} p(s \rightarrow b_j )$$

\noindent $\sharp b_i$ is the size of the basin $b_i$.

\noindent Now we can define a \textit{Local Optima Network} (LON)
$G=(S^*,E)$ as being the graph where the nodes
are the local optima, and there is an edge $e_{ij} \in E$ with
weight $w_{ij} = p(b_i \rightarrow b_j)$ between two nodes $i$ and
$j$ if $p(b_i \rightarrow b_j) > 0$.
Notice that since each maximum has its associated basin, $G$ also
describes the interconnection of basins.

\noindent According to our definition of edge weights, $w_{ij} = p(b_i
\rightarrow b_j)$ may be different than $w_{ji} = p(b_j \rightarrow
b_i)$. Thus, two weights are needed in general, and we have an
oriented transition graph. Clearly, different  move operators and thus different neighborhood structure
will induce different LONs.

\begin{comment}
\noindent Finally, the following two definitions are relevant to the discussion of the
boundary of basins.
The \textit{boundary} $B(b)$ of a basin of attraction $b$ can be
defined as the set of configurations within a basin that have at
least one neighbor's solution in another basin $b^{'}$.
Conversely, the \textit{interior} $I(b)$ of a basin is composed by
the configurations that have all their neighbors in the same basin.
Formally,
$$
B(b) = \lbrace s \in b \; | \: \exists \: b^{'} \not= b, \: \exists \: s^{'} \in
b^{'}, \: \exists \: e_{ss^{'}} \in E \rbrace,
$$
\vspace{-1cm}

$$I(b) = b-B(b)
$$
\end{comment}

\section{Analysis of the local optima network}
\label{sec:analysis}

\subsection{Experimental settings}
\label{sec:experimentalSetting}

In order to perform a statistical analysis, a sufficient number of instances have to be considered. Well-known benchmark instances
producing two distinct categories of QAP problems are those
of  Knowles and Corne~\cite{Knowles2003emo} which have been adapted and used here for the single-objective QAP. 

The first generator produces uniformly random instances where all flows and distances are integers sampled from uniform distributions in $\left[1,f_{max}\right]$ and $\left[1,d_{max}\right]$ respectively; this leads to the same kind of problem known in literature as \emph{Tai}\verb"nn"\emph{a}, being \verb"nn" the problem dimension~\cite{Taillard1995}\footnote{there, though, random values are uniformly distributed in $\left[0,99\right]$}. Distance matrix entries are, in both cases, the Euclidean distances between points in the plane.

The second generator permits to obtain clusters of $1$ to $K$ points that are uniformly distributed in small circular regions of radius $m$, with these regions distributed in a larger circle of radius $M$. In this case then, the flow entries are non-uniform random values, controlled by two parameters, $A$ and $B$, with $A < B$, and $B > 0$. Let $X$ be a random variable uniformly distributed in $\left[0,1\right]$, then a flow entry is given by integer rounding $10^{(B-A)*X + A}$.
When the values of $A$ is negative, the flow matrix is sparse and non-zero entries are non-uniformly distributed.
This procedure, detailed in~\cite{Knowles2003emo}, follows the one introduced by Taillard~\cite{Taillard1995} and produces random instances of type \emph{Tai}\verb"nn"\emph{b} which have the so called ``real-like'' structure.

For the following analysis, 30 random uniform and 30 random real-like instances have been generated for each problem dimension in $\{5,...,10\}$; as for the distance matrix, values of 100 for $d_{max}$ in the first case and of $(0,1,100)$ for $(M,K,m)$ in the second have been chosen\footnote{with this choice for units position distribution, real-like instances really differ from uniform one only from the flow matrix point of view.}; as for the flow matrix, the parameters used have been $f_{max}=100$ for the uniform instances, $A=-10$ and $B=5$ for the others. The latter choice, in particular, results in a flow matrix with roughly two thirds of out-diagonal zeros in the real-like case.

\subsection{General network features}

\subsubsection{Nodes and Edges}

\begin{figure}[ht]
\begin{center}
 \includegraphics[width=0.40\textwidth]{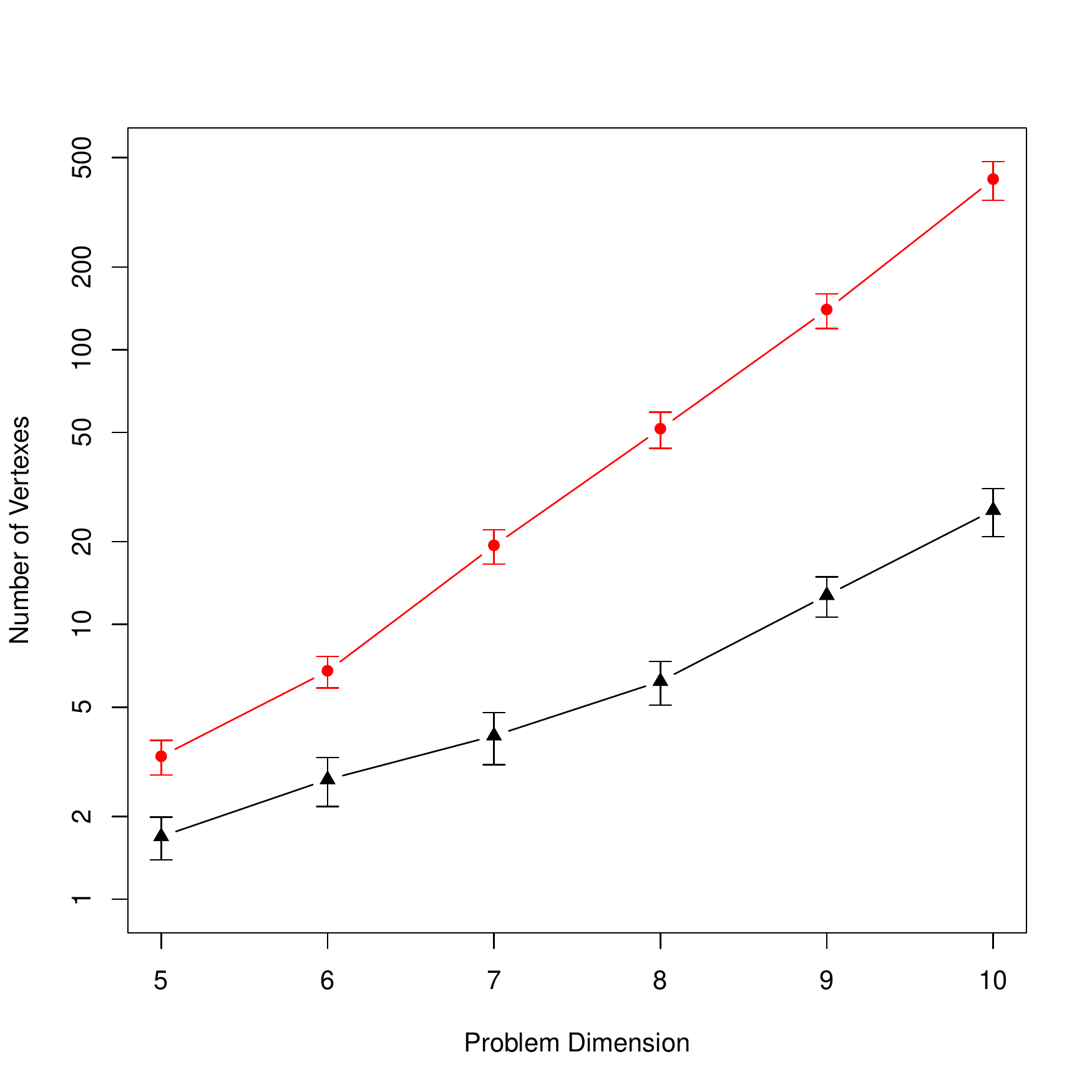}
 %\hspace{5pt}
% \vspace{-5pt}
 \includegraphics[width=0.40\textwidth]{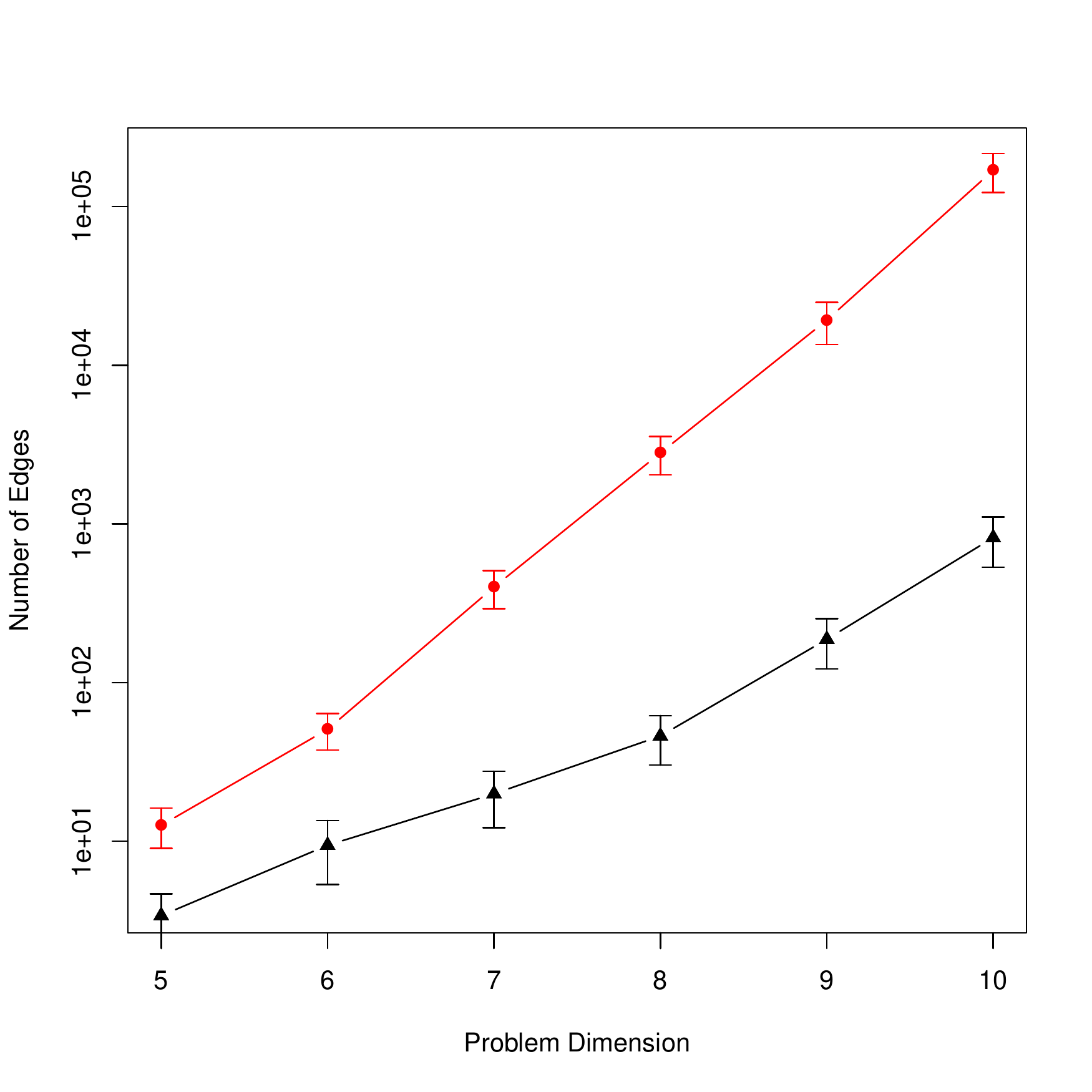}
 %\vspace{-5pt}
 \end{center}
 \vspace{-0.3cm}
 \caption{Average number of nodes (top) and edges (bottom) on log-lin scale. Triangular points correspond to real-like problems, rounded points to uniform ones; bars show 95\% Wald C.I. on the means; for each problem dimension, averages from 30 independent and randomly generated instances are shown.}
\label{fig:nv}
\end{figure}

Figure~\ref{fig:nv} (top) reports, for each problem dimension, the average number of nodes found in the local optima networks. The search difficulty of these landscape is expected to increase with the number of local optima, and this value grows exponentially with the problem dimension. Real-like instances, though, result in much smaller networks (small number of vertexes); the size difference between the two classes of QAP also grows almost exponentially with the problem dimension.

Figure~\ref{fig:nv} (bottom) shows a similar growth  for the average number of edges. Indeed, the graphs are almost fully connected, i.e. the number of oriented edges is close to the squared number of nodes.

\enlargethispage{\baselineskip}

\subsubsection{Basins of attraction}

\begin{figure}[h!]
\begin{center}
 \includegraphics[width=0.40\textwidth]{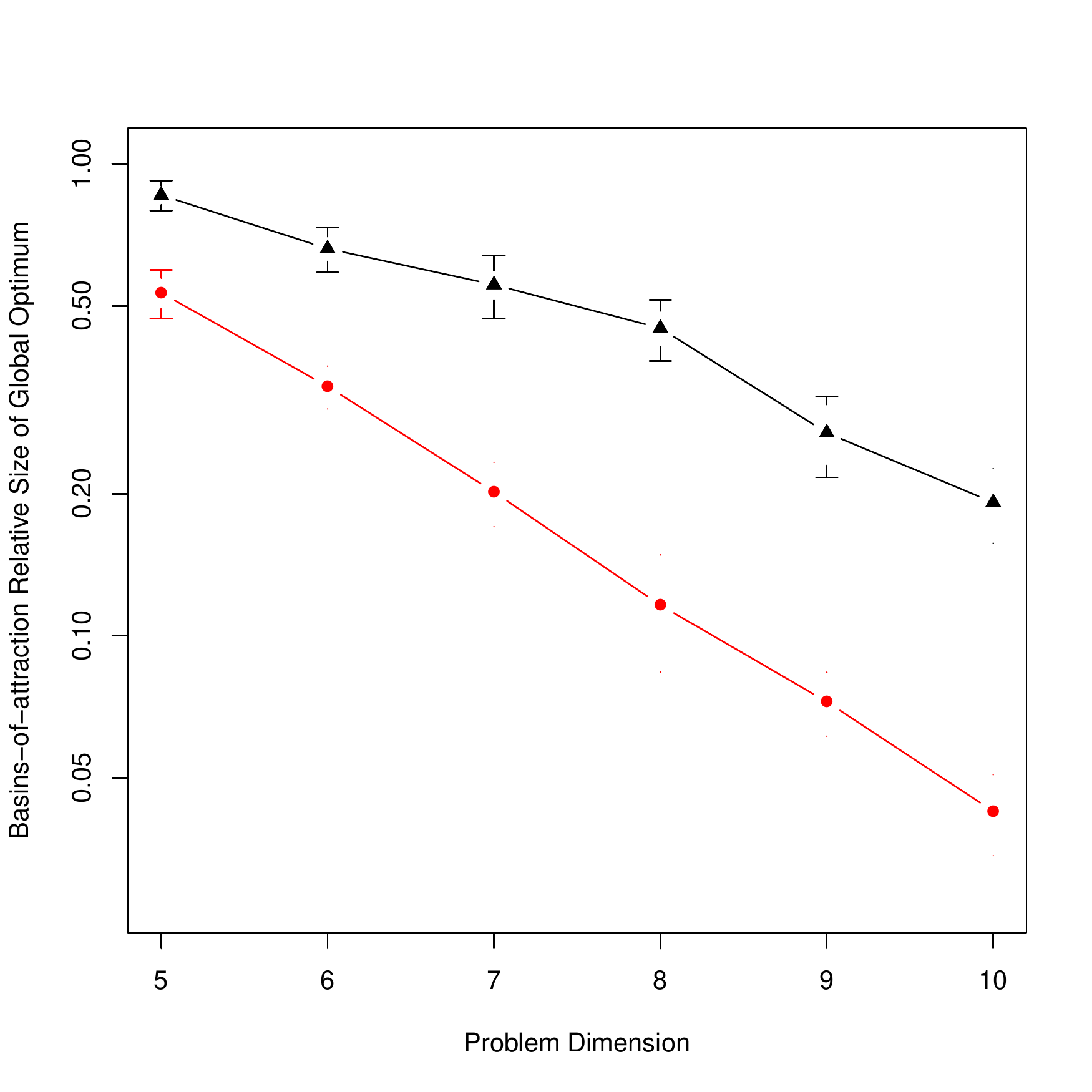}
  \vspace{-0.3cm}
 \caption{Average relative size of the global optimum basin-of-attraction on log-lin scale. Triangular points correspond to real-like problems, rounded points to uniform ones; bars show 95\% Wald C.I. on the means; for each problem dimension, averages from 30 independent and randomly generated instances are shown.}
\label{fig:bas}
\end{center}
\end{figure}

Figure~\ref{fig:bas} depicts the average size of the basin of attraction of the global optimum divided by the size of the search space. This value decreases exponentially with the problem dimension for both considered classes of QAP instances. The real-like instances present larger global optimum basins, which can be explained by their smaller local optima networks (this suggested explanation is further elaborated below). The relative size of the global optimum basin gives the probability of finding the best solution with a hill-climbing algorithm from a random starting point. The exponential decrease confirms that the higher  the problem dimension, the lower  the probability for a stochastic search algorithm to locate the basin of attraction of the global minimum. Considering the separation between the curves in fig.~\ref{fig:bas}, it looks surprisingly easier to solve exactly a real-like instance rather than a uniform one.

\begin{figure}[h!]
\begin{center}
 \includegraphics[width=0.40\textwidth]{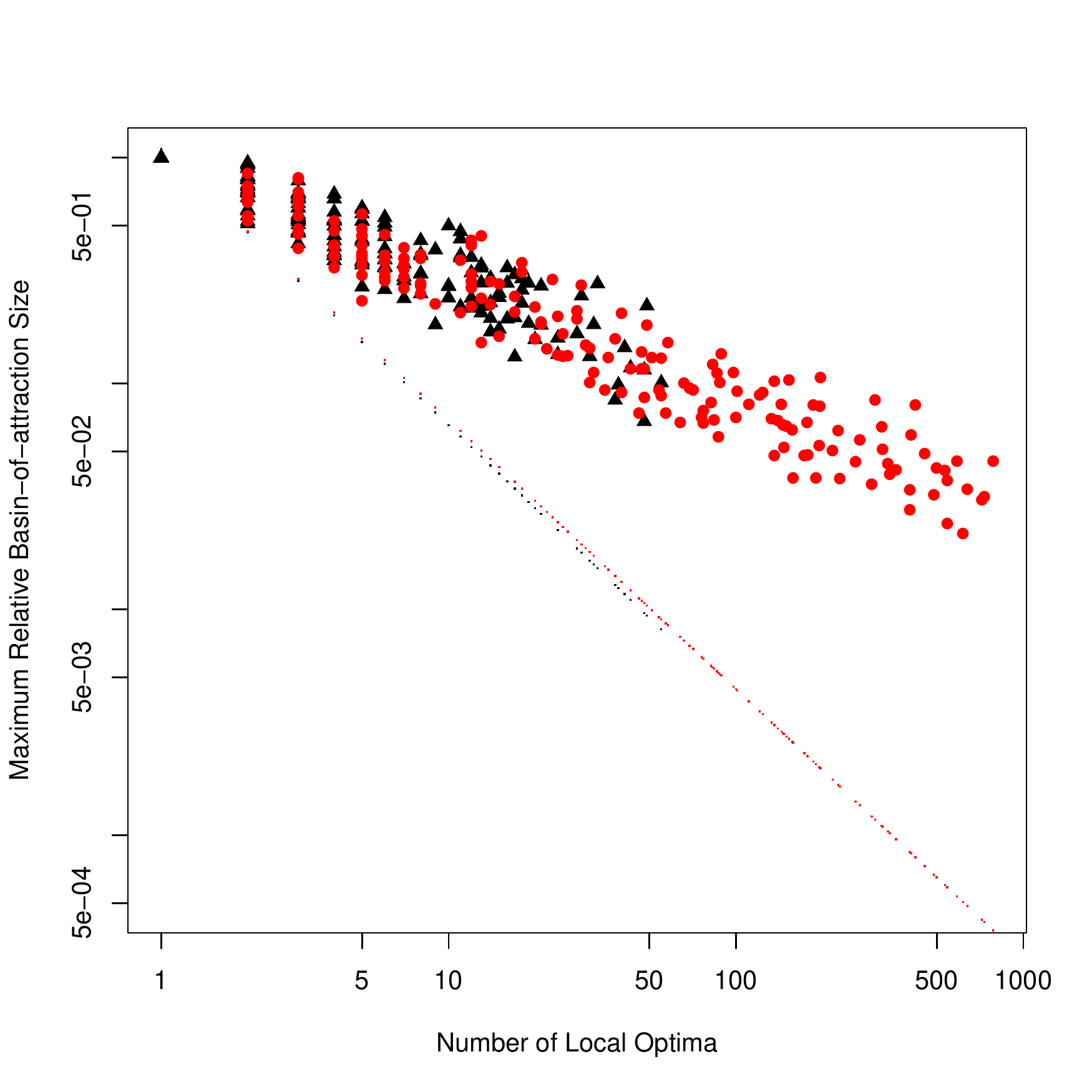}
 \includegraphics[width=0.40\textwidth]{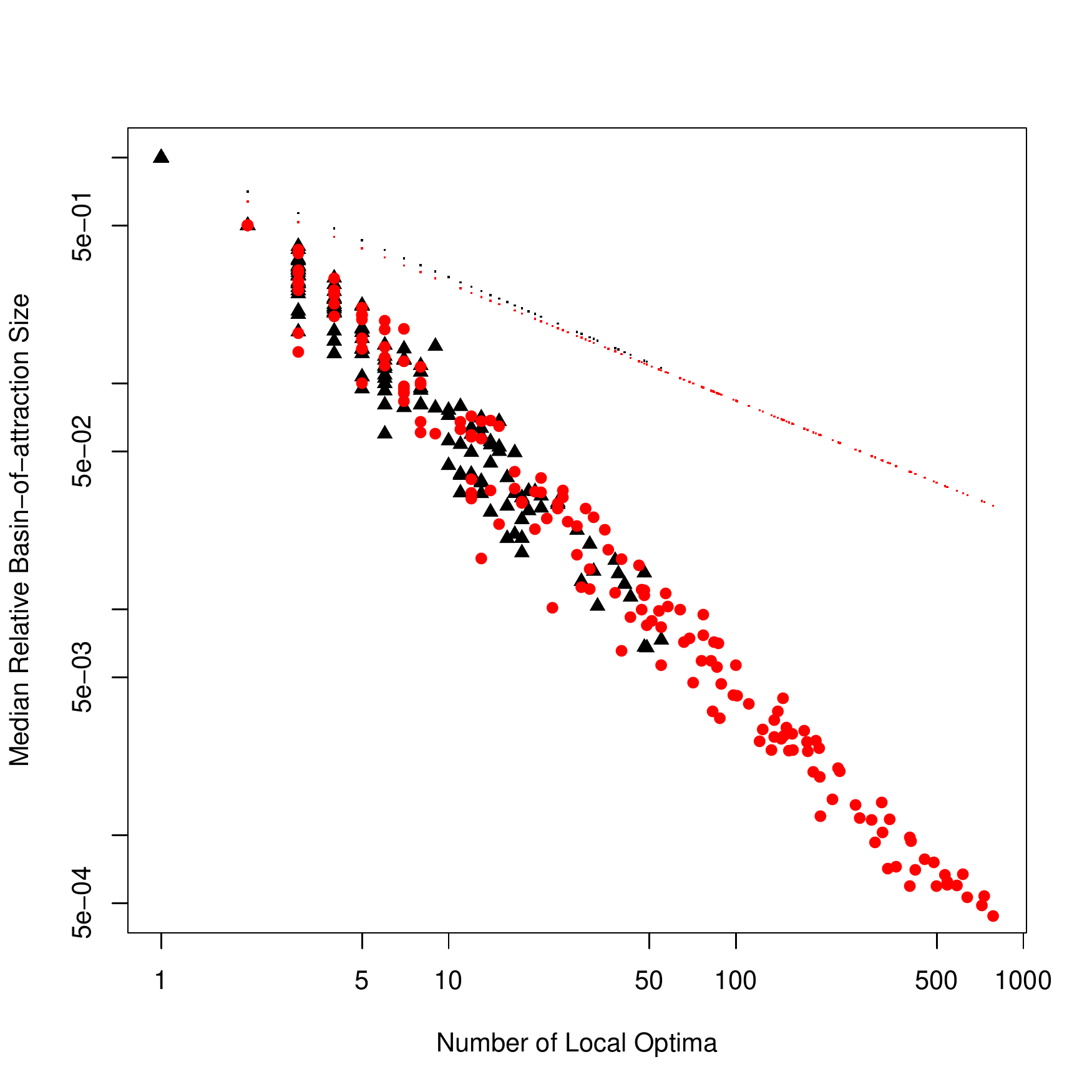}
  \vspace{-0.3cm}
 \caption{Relative size of the largest (top) and of the median (bottom) basin of attraction vs number of nodes. Triangular points correspond to real-like instances, rounded points to uniform ones. Each figure reports in dotted form the regression lines of the other.}
\label{fig:basrel}
\end{center}
\end{figure}

The distribution of basins sizes is very asymmetrical, thus the median and maximum sizes are used as representatives\footnote{the average basins size, equal to the number of possible configuration in the search space divided by the number of local optima, is not informative}.  These statistics are divided by the search space size and plotted against the number of local optima (figure~\ref{fig:basrel}) in order to convey a view independent of problem dimension and LON cardinality.

The median and maximum basin sizes  follow a power-law relationship with the number of local optima. The difference between the two values tends to diverge exponentially with the LON size.  This suggests that the landscapes are characterized by many small basins and few larger ones. The correlation coefficients of the real-like and uniform instances are similar. Both instance classes seem to have the same size distribution with respect to the local optima network cardinality. This fact, coupled with the different number of local optima but a similar distribution of basin sizes, can explain how real-like instances have larger global optimum basin compared to uniform instances of the same problem dimension.

\begin{figure}[h!]
\begin{center}
 \includegraphics[width=0.40\textwidth]{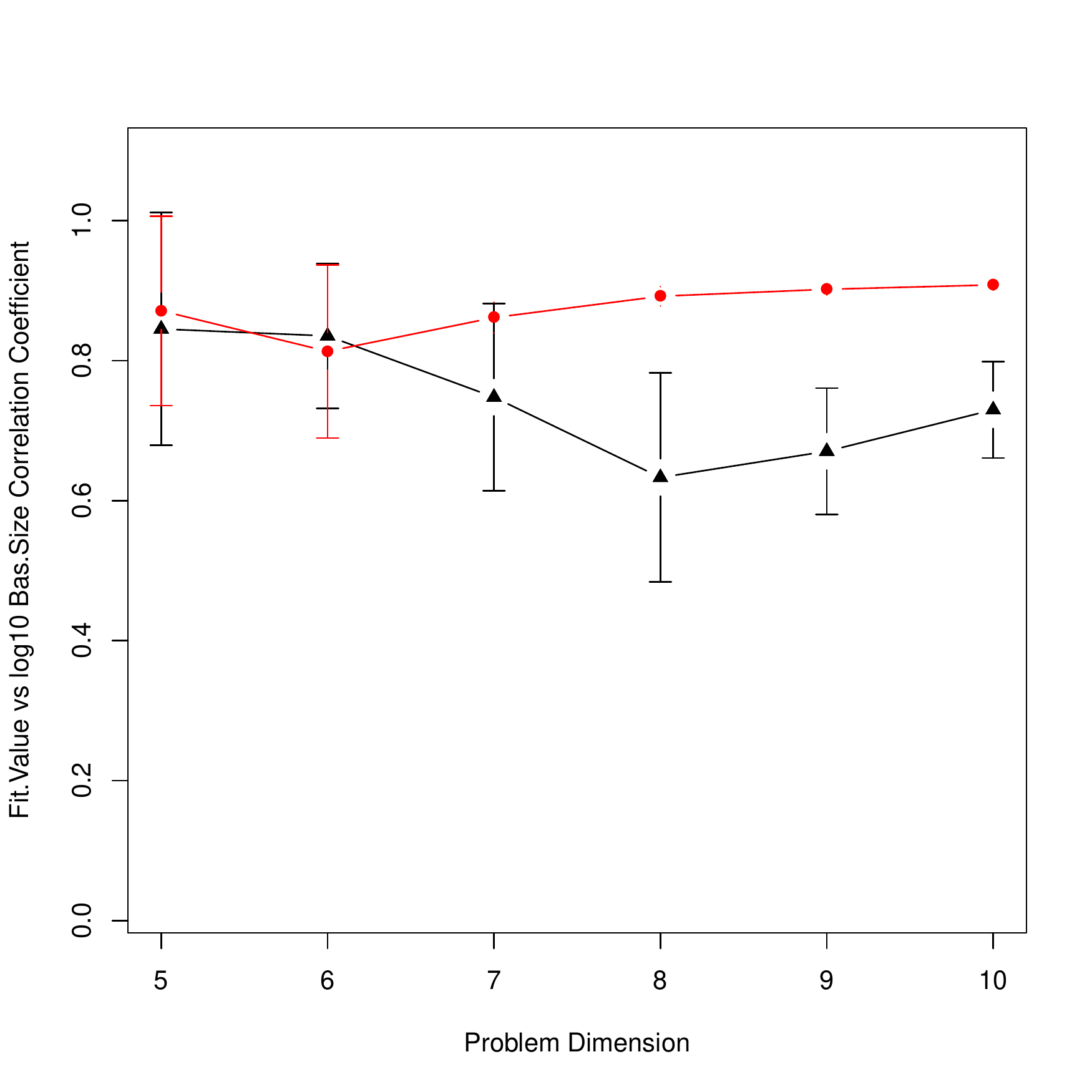}
  \vspace{-0.3cm}
 \caption{Average Fit.Value vs Bas.Size Correlation Coefficient. Triangular points correspond to real-like problems, rounded points to uniform ones; bars show 95\% Wald C.I. on the means; for each problem dimension, averages from 30 independent and randomly generated instances are shown.}
\label{fig:corrfitbas}
\end{center}
\end{figure}

Figure~\ref{fig:corrfitbas} plots the correlation coefficients between the logarithm of local optima basin sizes and their fitness value. There is a strong positive correlation. In other words, generally the better the fitness value of an optimum, the wider its basin of attraction. It is worth noticing, however, that the relative size of the global optimum (to the search space dimension), decreases exponentially as the problem size increases (see fig.\ref{fig:bas}). Real-like and uniform instances show a similar behavior but the  former present higher variability and slightly lower correlation figures.

\begin{figure}[h!]
\begin{center}
 \includegraphics[width=0.40\textwidth]{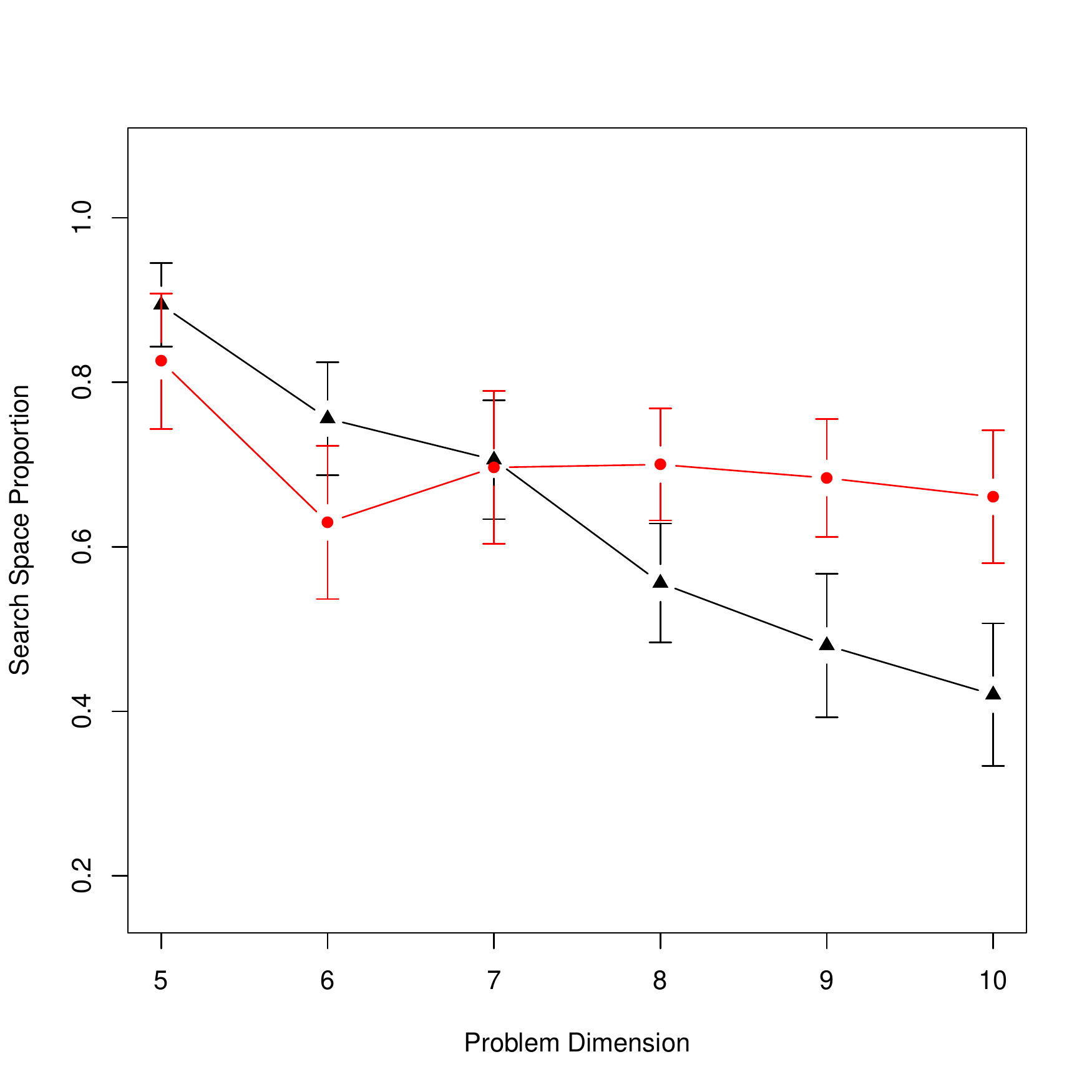}
  \vspace{-0.3cm}
 \caption{Proportion of search space whose solutions climb to a fitness value within 5\% from the global best value.
Triangular points correspond to real-like problems, rounded points to uniform ones; for each problem dimension, averages from 30 independent and randomly generated instances are shown.}
 \label{fig:fitd}
\end{center}
\end{figure}

From what has been studied, real-like instances are easier to solve exactly using heuristic search. However, Merz and Freisleben \cite{merz2000fitness} have shown that the quality of local optima decreases when there are few off-diagonal zeros in the flow matrix: the cost contributions to the fitness value in eq.\ref{eq:fitness} are in that case more interdependent, as there was a higher epistasis. Thus it should be easier to find a sub-optimal solution for a uniform instance than for a real-like one. To confirm this result in another way, figure~\ref{fig:fitd} shows the proportion of solutions from whose a best improvement hill-climbing conducts to a local optima within a $5\%$ value from the global optimum cost. As problem size grows, sub-optimal solutions are distinctively easier to reach in the uniform case, i.e. uniform instances are easier to be solved approximatively. This also agrees with the fact that for large enough instances, the cost ratio between the best and the worst solution has been proved to converge to one in the random uniform case~\cite{krokhmal2009random}.

\subsubsection{Transition probabilities}

\begin{figure}[h!]
\begin{center}
 \includegraphics[width=0.40\textwidth]{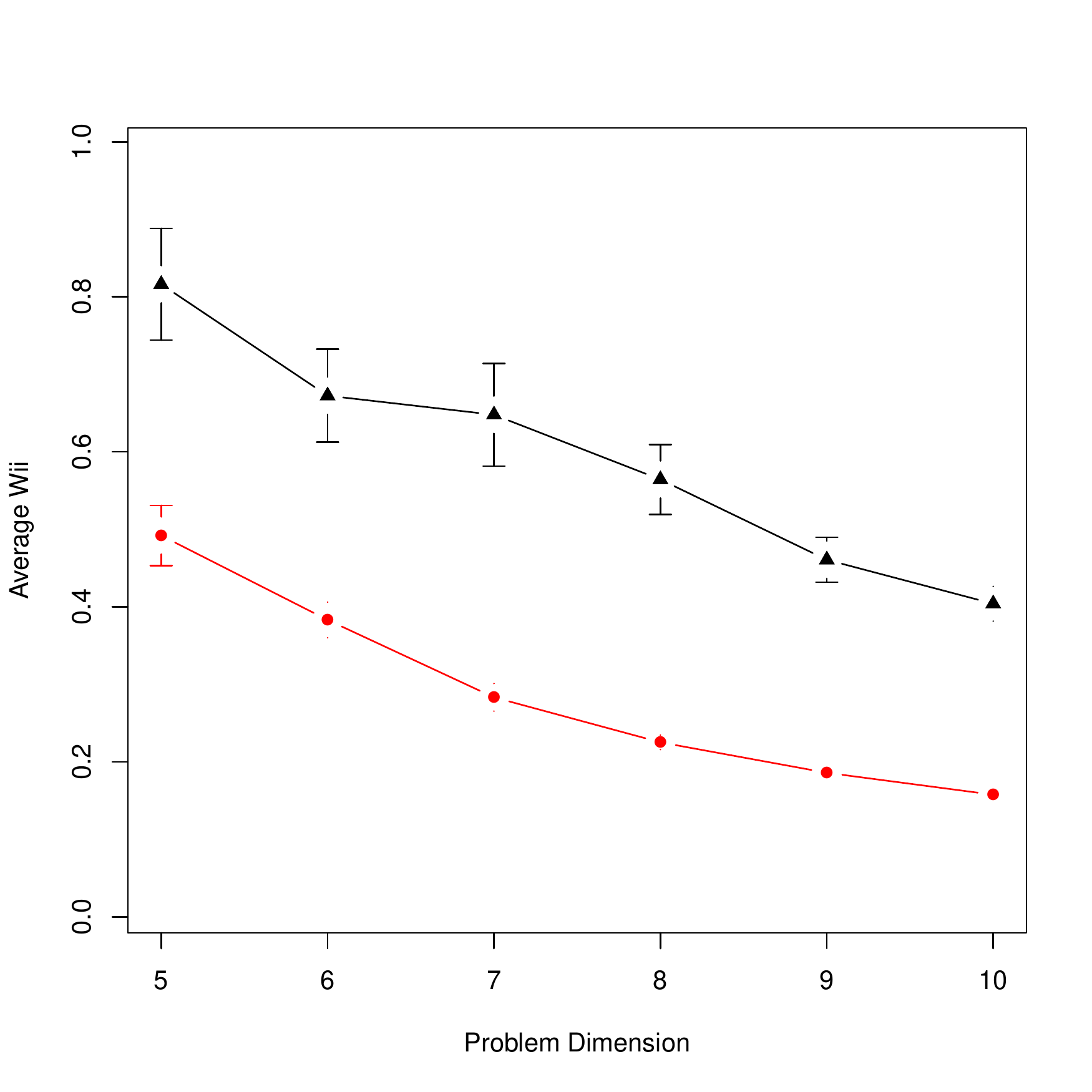}
 \includegraphics[width=0.40\textwidth]{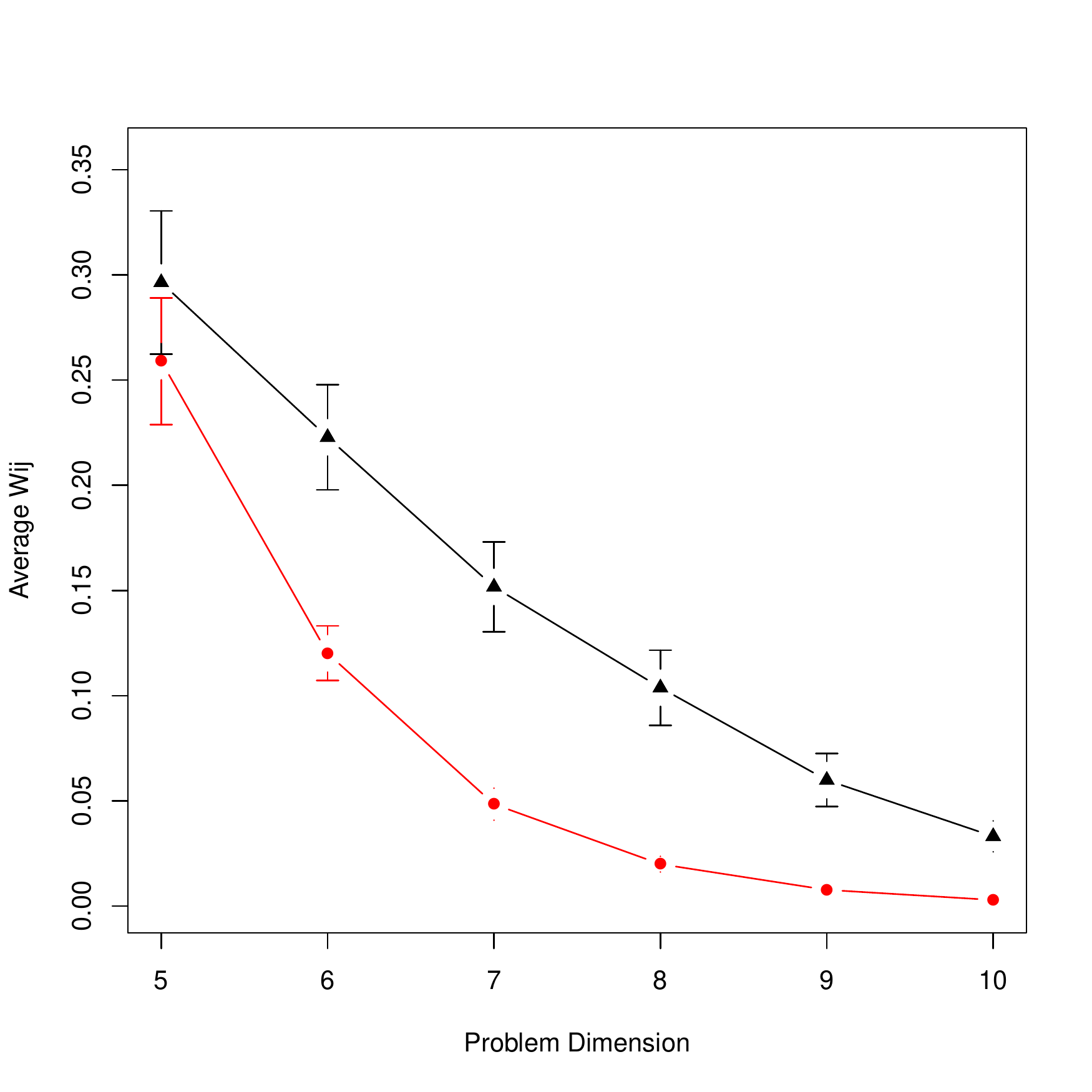}
  \vspace{-0.3cm}
 \caption{Average weights $w_{ii}$ for self-loop (top) and $w_{ij}$ for out-going links (bottom). Triangular points correspond to real-like instances, rounded points to uniform ones; bars show 95\% Wald C.I. on the means; for each problem dimension, averages from 30 independent and randomly generated instances are shown.}
\label{fig:wii}
\end{center}
\end{figure}

Figure~\ref{fig:wii} (top) reports for each problem dimension the average weight $w_{ii}$ of self-loop edges. These values represent the one-step probability of remaining in the same basin  after a random move. The higher values observed for real-like instances are related to their fewer optima but bigger basins of attraction. However, the trend is generally decreasing with the problem dimension. This is another confirmation that basins are shrinking with respect to their relative size.

A similar behavior characterizes  the average weight $w_{ij}$ of the outgoing links from each vertex $i$, as figure~\ref{fig:wii} (bottom) reports. Since  $j \neq i$, these weights represent the probability of reaching the basin of attraction of one of the neighboring local optima. These probabilities decrease with the problem dimension. The difference between the two classes of QAP could be explained here by their different LON size.

A clear difference in magnitude  between $w_{ii}$ and $w_{ij}$ can be observed. This means that, after a move operation,  it is more likely to remain in the same basin than to reach another basin. Moreover, the decreasing trend with the problem dimension  is stronger for $w_{ij}$ than for $w_{ii}$, especially for the uniform QAP (whose LON grows faster with the problem dimension). Therefore, even for these small instances, the probability of reaching a particular neighboring basin becomes rapidly smaller than the probability of staying in the same basin, by an order of magnitude.

\subsubsection{Weighted connectivity}

In weighted networks, the degree of nodes is extended by defining the node \textit{strength} $s_i$ as the sum of all the weights $w_{ij}$ of the links attached to it. This value gathers information from both the connectivity of a node and  the importance of its links~\cite{bart05}. In our definition, the out-going  weights always sum up to 1, so the out-going strength is just $1-w_{ii}$. Therefore, a study of $s_i$ associated to the in-coming links would be more informative.

\begin{figure}[h!]
\begin{center}
 \includegraphics[width=0.40\textwidth]{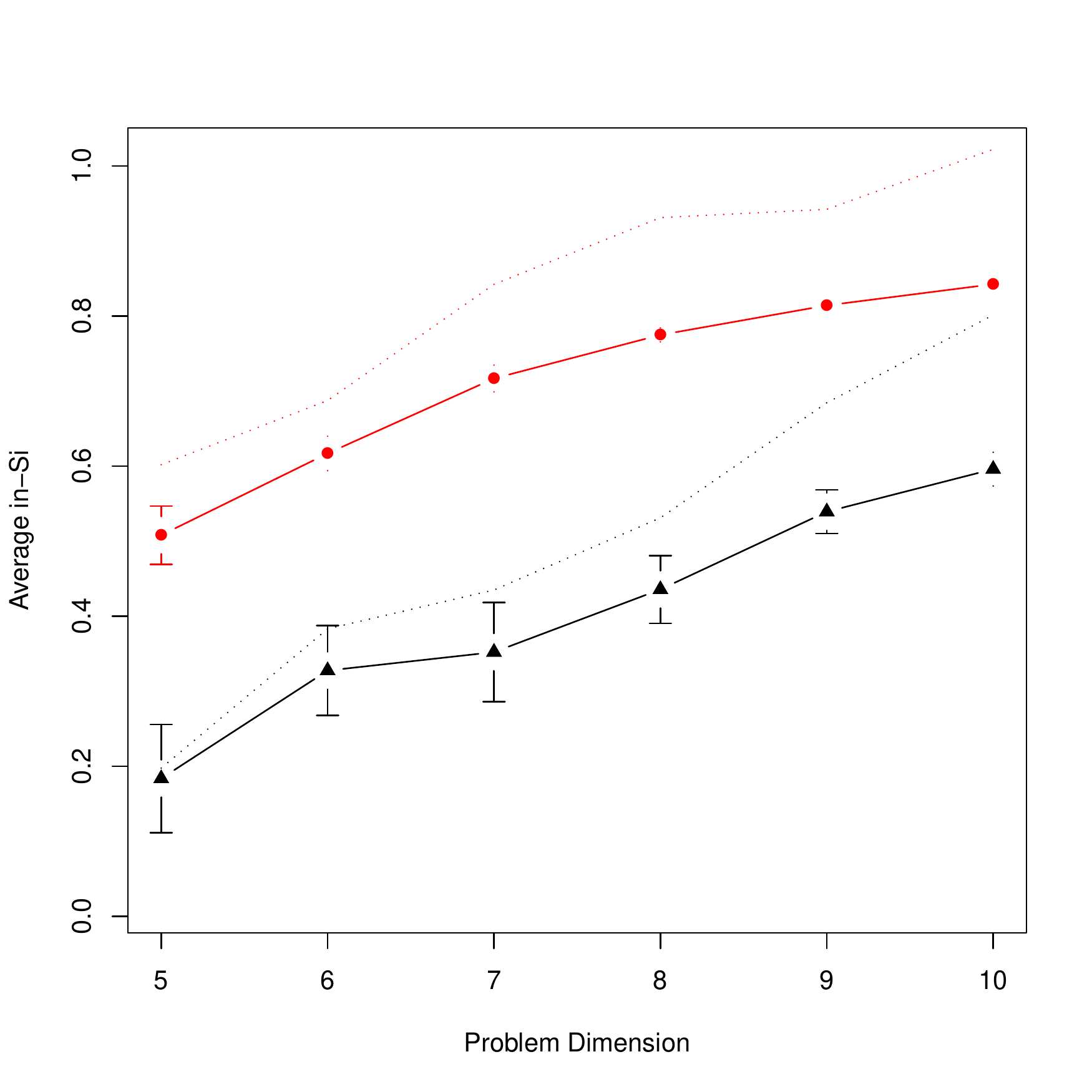}
  \vspace{-0.3cm}
 \caption{Average vertexes in-coming strength. Triangular points correspond to real-like problems, rounded points to uniform ones; bars show 95\% Wald C.I. on the means; dotted lines report the mean in-coming degree multiplied by the mean edge weight; for each problem dimension, averages from 30 independent and randomly generated instances are shown.}
\label{fig:sin}
\end{center}
\end{figure}

Figure~\ref{fig:sin} reports the average vertex strength for connections entering the considered basin. The increase with the problem dimension and the separation between the two QAP classes should come as no surprise since strength is related to vertex degree (which is higher for bigger LONs, given that they are almost complete). In particular, if  the distribution of weights were independent of topology, the strength would be simply equal to the vertex degree multiplied by the mean edge weight. These expected curves are plotted in dotted form and, although there is some clear correlation, the actual strengths are distinctively lower.

\begin{figure}[h!]
\begin{center}
 \includegraphics[width=0.40\textwidth]{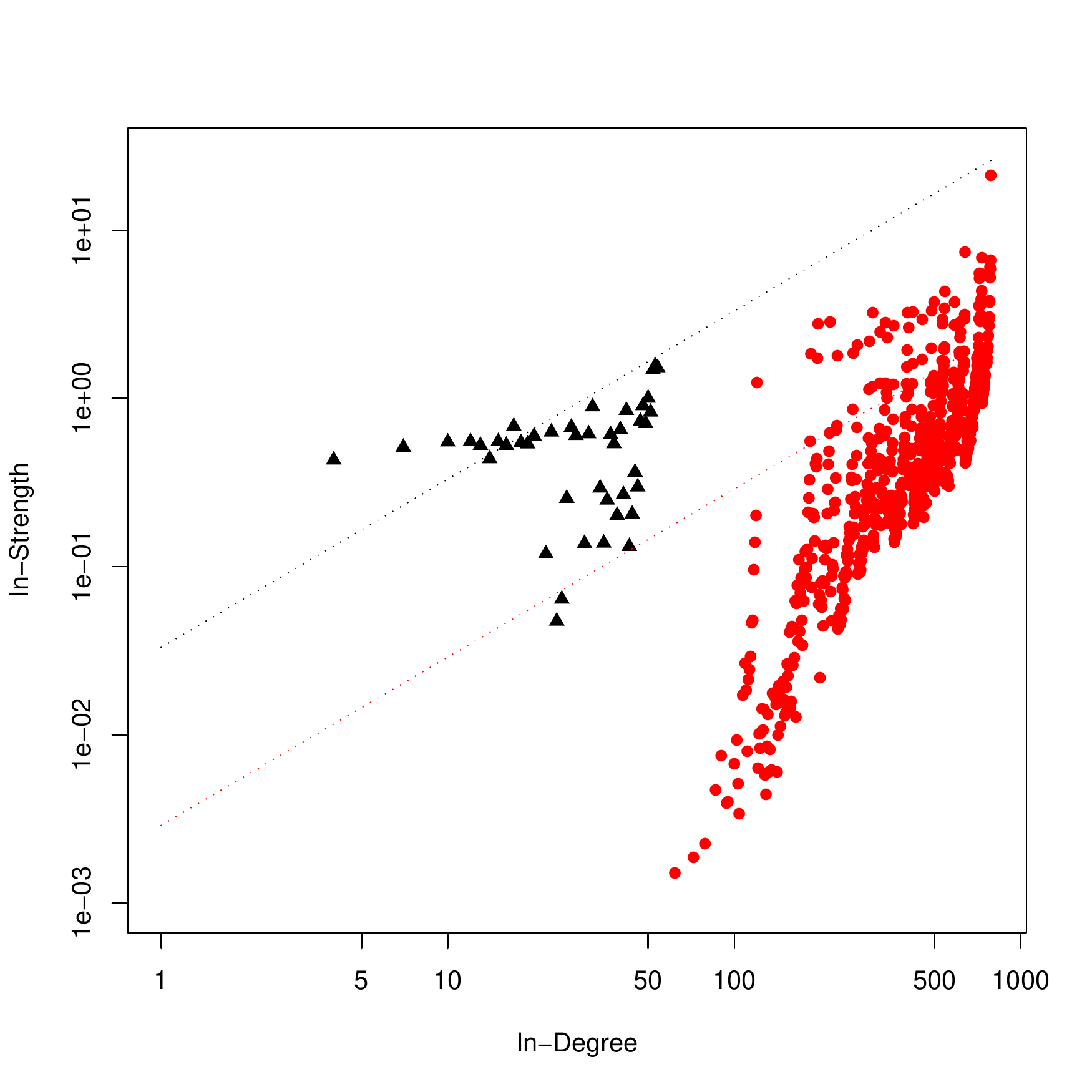}
  \vspace{-0.3cm}
 \caption{Aggregated average of in-strength to vertex in-degree on log-log scale. Triangular points correspond to real-like problems, rounded points to uniform ones; all 30 independent and randomly generated instances of problem dimension 10 are shown. Dotted lines report the  $s=\overline{w}*k$ relation.}
\label{fig:sindeg}
\end{center}
\end{figure}

 Figure~\ref{fig:sindeg} shows the aggregated average of $s_i$  with respect to the vertex in-degree $k_i$  for all the instances of dimension 10. As the figure suggests, a fit with the law $s=\overline{w}*k$ does not hold. Given the high connectivity of these LONs, the degree values are all close to the maximum possible for each network.  Thus, the strength of vertexes must grow much faster than their degree in order to give the averages seen in fig.~\ref{fig:sin}. It can be observed that, for uniform QAP instances, the scatter plot of fig.~\ref{fig:sindeg}  is constituted by several curves that  are almost vertical even in log-log scale. Real-like instances, on the contrary, have so smaller and  densely connected LONs that an analogue behavior can not be spotted. However, for both  instance classes,  fig.~\ref{fig:sindeg} clearly suggests that the strength is far from being simply proportional to the vertex degree. Therefore,  we have a confirmation that the distribution of weights, as we have defined them, strongly depends on the network topology.

\begin{figure}[h!]
\begin{center}
 \includegraphics[width=0.40\textwidth]{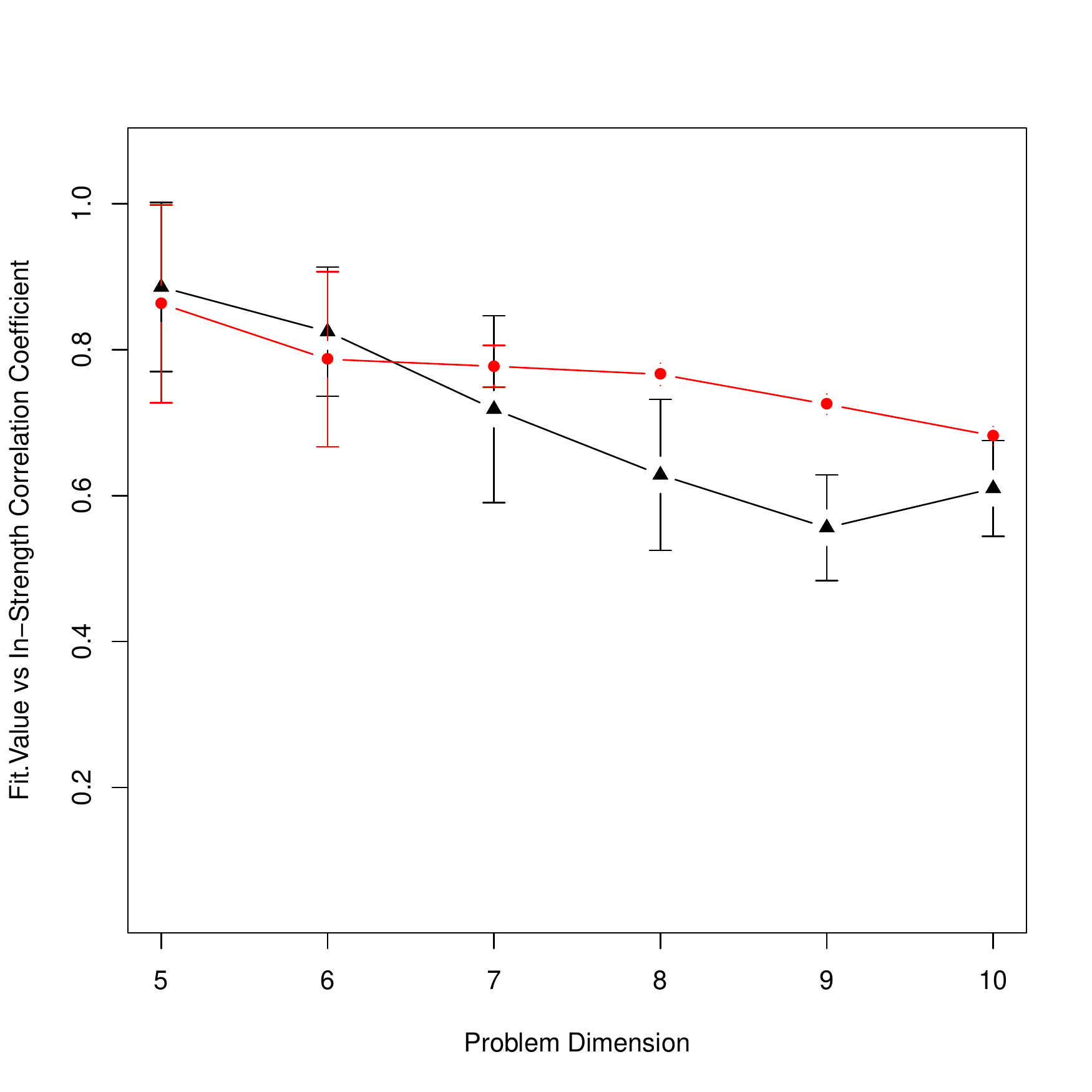}
  \vspace{-0.3cm}
 \caption{Average Fit.Value vs In-Strength Correlation Coefficient. Triangular points correspond to real-like problems, rounded points to uniform ones; bars show 95\% Wald C.I. on the means; for each problem dimension, averages from 30 independent and randomly generated instances are shown.}
\label{fig:corrfitsin}
\end{center}
\end{figure}

Figure~\ref{fig:corrfitsin} reports the correlation coefficients between the in-coming strength of a vertex and the fitness value of its local optimum. The correlation is positive and strong, which suggests that  basins with high fitness generally have more heavy weighted connectivity. Our explanation to this observation is the following:  with our definition of transition probabilities, basin sizes have an influence on weight values. Also,  there is a high correlation between the logarithmic size of a basin and its fitness value  (see fig.~\ref{fig:corrfitbas}).The same considerations made there still hold here, with the difference that the link between strength and fitness seems less tight as the size of problem grows. This could add to the search difficulty.

\subsection{Advanced network features}

\subsubsection{Transitivity}
Transitivity measures the probability that the adjacent vertexes of a vertex are connected~\cite{wasserman1994social}, this feature is measured with the so called {\em clustering coefficient}. The traditional definition of this coefficient does not consider weights,  thus, it has been extended in several ways to a ~\textit{weighted clustering coefficient}~\cite{bart05}. Since our studied LONs from the QAP instances are close to be complete graphs, we selected the simplest definition. All the instances were found to have a transitivity value higher than $0.90$: real-like instances have a value above $0.99$, whereas uniform instances present a slight decrease with respect to the size of the problem. We also observed that this measure has a lower variability when compared to the other networks statistics. Thus, a very high clustering coefficient appears to be an instance independent characteristic of QAP with the given definition of LON.

\subsubsection{Disparity}

Another network statistic, which measures how heterogeneous are the contributions of the edges of a node $i$ to its strength $s_i$, is \textit{disparity}~\cite{bart05}. Disparity could be defined as $Y_{2}(i) = \sum_{j \not= i} ( \frac{w_{ij}}{s_i} )^2$ and could be averaged over the nodes with the same degree $k$. If all the weights $w_{ij}$ are close to $s_i/k_i$, then $Y_2(i) \approx 1/k_i$ for nodes of degree $k_i$.

\begin{figure}[h!]
\begin{center}
 \includegraphics[width=0.40\textwidth]{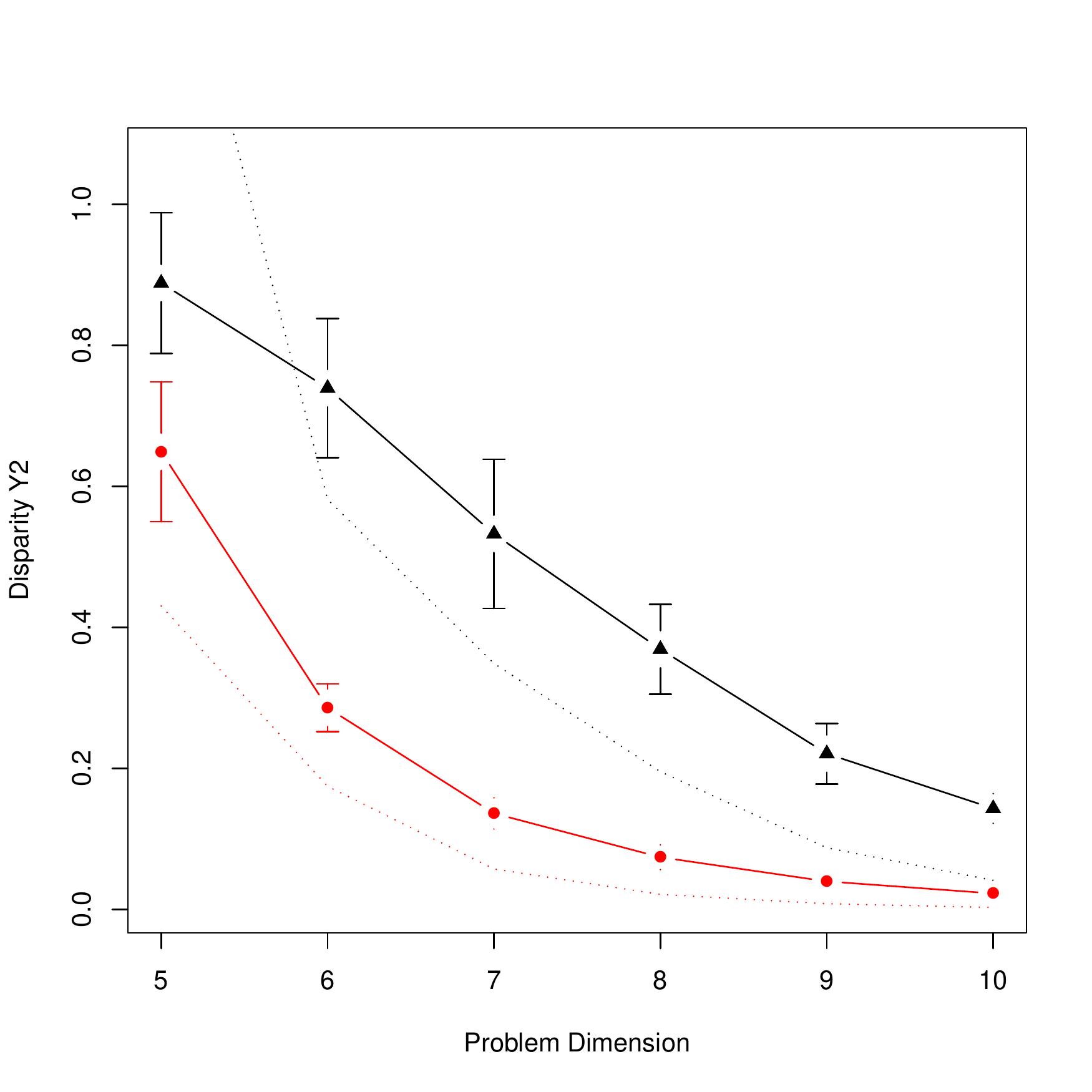}
  \vspace{-0.3cm}
 \caption{Average disparity coefficient. Triangular points correspond to real-like instances, rounded points to uniform ones; bars show 95\% Wald C.I. on the means; for each problem dimension, averages from 30 independent and randomly generated instances are shown.}
\label{fig:y2}
\end{center}
\end{figure}

Figure~\ref{fig:y2} reports the simple mean of disparity coefficient averaged on all instances of both classes with respect to problem dimension. The decreasing trend could render the fact that as the size of the problem rises, the out-going transition to different optima neighbors tend to become equally probable and the search becomes more random. That could be more evident for uniform instances whose LONs have higher cardinality. Real-like instances, actually, appear to maintain a disparity value less close to the random curve of $1/\overline{k}$.

\begin{figure}[h!]
\begin{center}
 \includegraphics[width=0.40\textwidth]{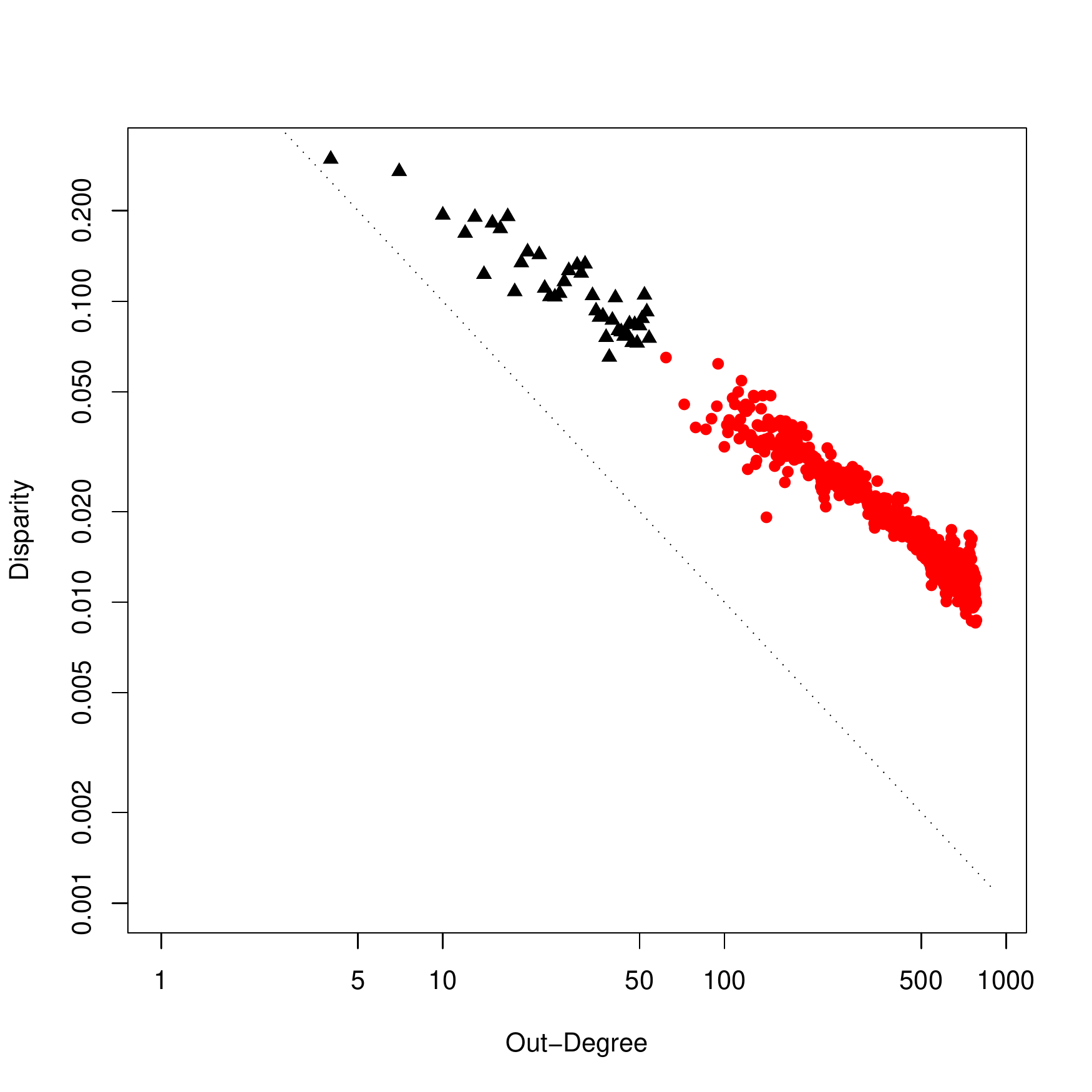}
  \vspace{-0.3cm}
 \caption{Aggregated average of disparity coefficient to vertex out-degree. Triangular points correspond to real-like instances, rounded points to uniform ones; all 30 independent and randomly generated instances for problem size 10 are shown. Dotted lines report the inverse of the  out-going degree.}
\label{fig:y2z}
\end{center}
\end{figure}

In figure~\ref{fig:y2z} the aggregate average of $Y_{2}(i)$ to $k_i$ is plotted on double logarithmic scale. Here just instances of problem dimension 10 have been considered. Disparity as a function of the node out-degree seems to follow a power-law, but as seen in fig.~\ref{fig:y2} that law is not the simple $1/k_i$. Thus it can be observed that, even if the weights $w_{ij}$ are not all equal to $s_i/k_i$, it remains difficult to spot an out-going connection whose probability dominates the others, except for really small problem instances\footnote{the one connection who really could rise disparity figures is the self-loop, but that has to be excluded by definition from the calculation of $Y_2$}.
According to the disparity measure, the real-like instances are not more difficult than uniform instances. No direction is pointed out by the weights distribution, and it must be considered to design efficient heuristics for QAP.

\subsubsection{Shortest paths}

\begin{figure}[h!]
\begin{center}
 \includegraphics[width=0.40\textwidth]{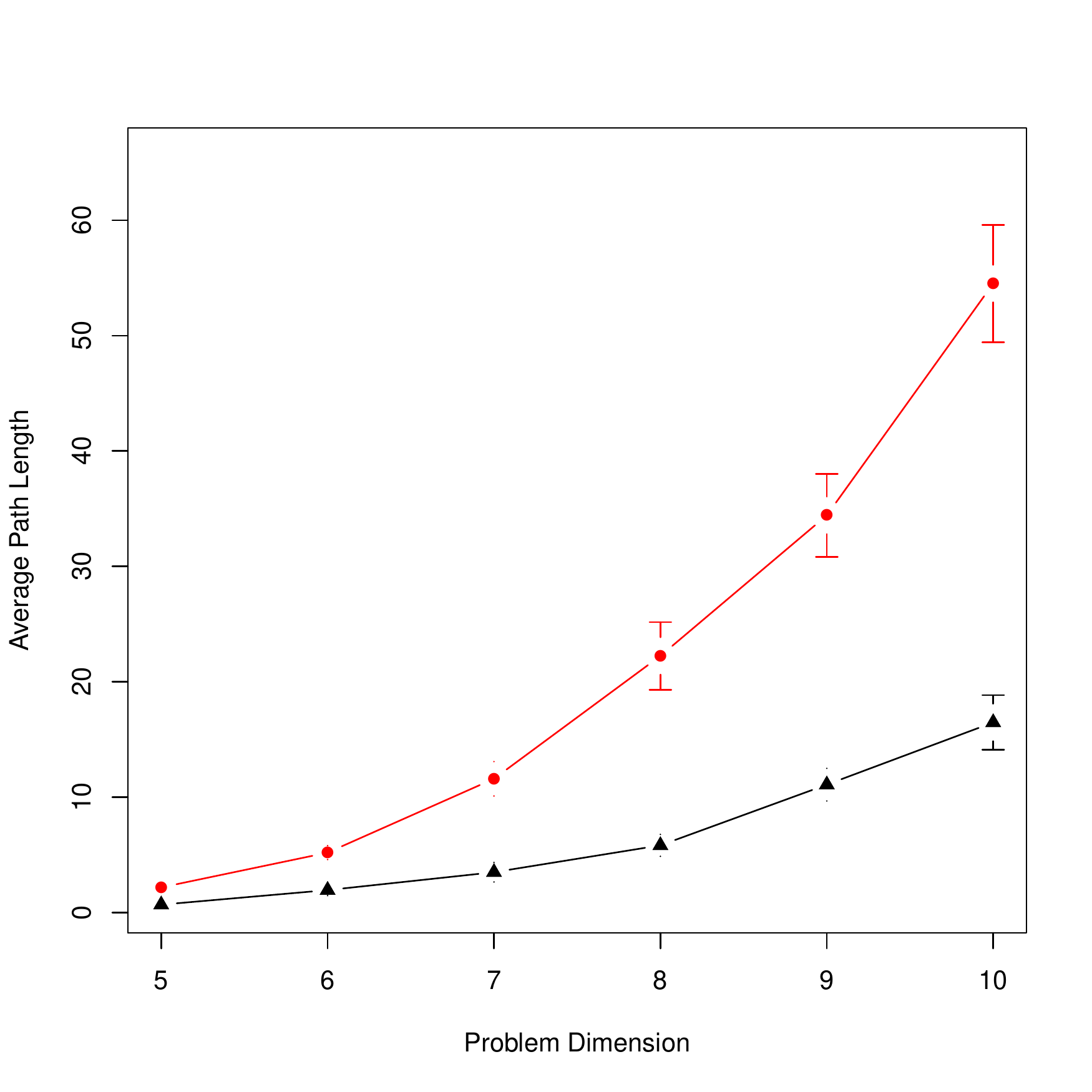}
 \includegraphics[width=0.40\textwidth]{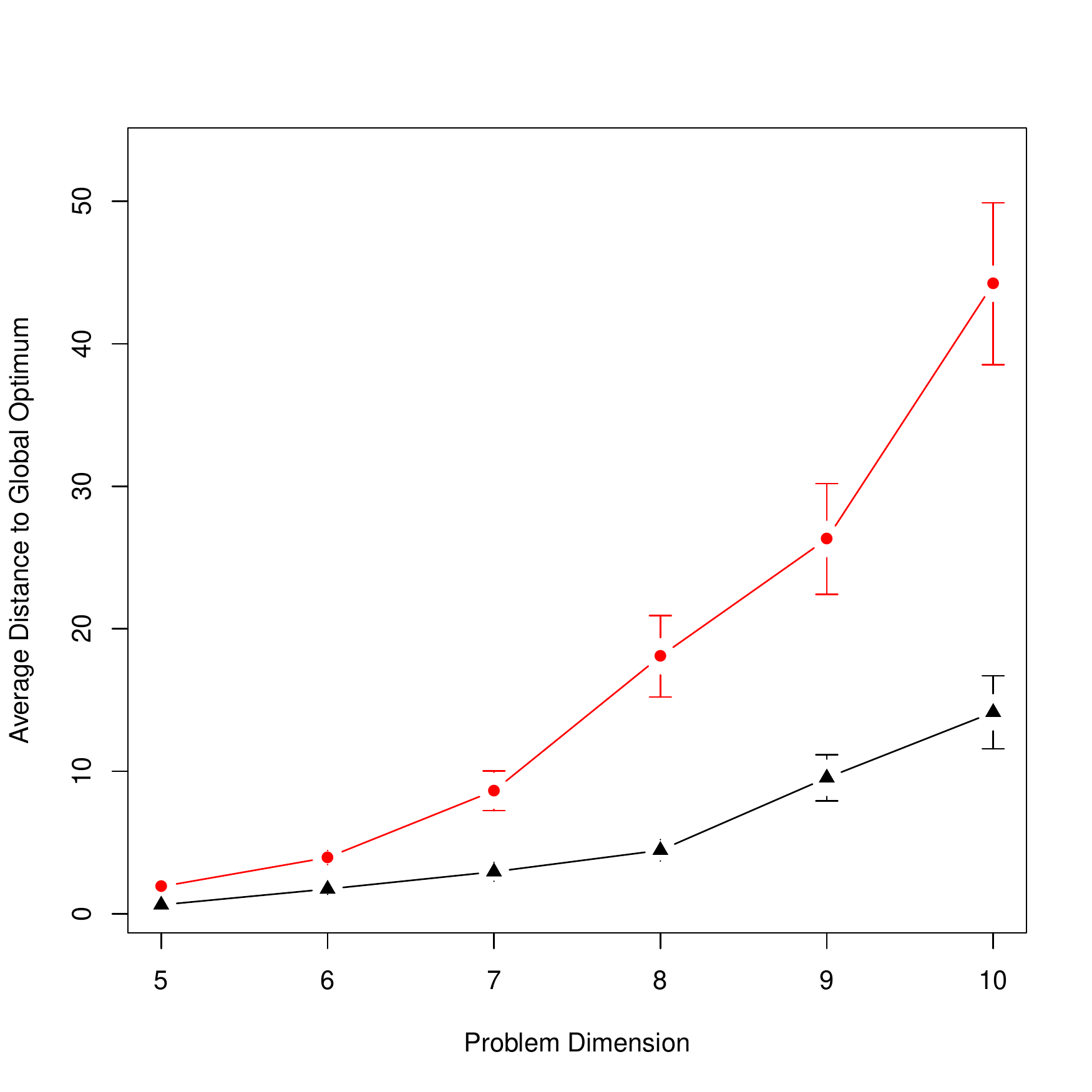}
  \vspace{-0.3cm}
 \caption{Average path length (top) average shortest path to global optimum (bottom). Triangular points correspond to real-like problems, rounded points to uniform ones; bars show 95\% Wald C.I. on the means; for each problem dimension, averages from 30 independent and randomly generated instances are shown.}
\label{fig:lv}
\end{center}
\end{figure}

A distance between two neighboring local optima $i$ and $j$  can be computed as the inverse of the transition probability between them: $1/w_{ij}$. This value can be interpreted as the expected number of random moves needed to hop from basin  $i$ to basin f $j$. The average path length can then be calculated  as the average of all the shortest paths between any two nodes (see figure~\ref{fig:lv} (top)).

Figure~\ref{fig:lv} (bottom),  reports a related measure, namely the mean shortest distance from each node to the global optimum. This metric can be more interesting from the point of view of a stochastic local search heuristic trying to solve the considered QAP instance. The clear trend is that this path, as any other, increases with the problem size. Values are noticeable higher for  the uniform instances, which have a larger number  of local optima than the real-like instance for  the same problem dimension. The figures confirm that the search difficulty increases with the domain size and  the ruggedness of the fitness landscape (\textit{i.e.} the number of local optima).

\section{Discussion and Conclusions}
\label{sec:discs}

We have used the recently proposed {\em Local Optima Network  (LON)} model to analyze the landscape of the well-known Quadratic Assignment Problem (QAP). Two types of instances: uniform and real-like, were  analyzed and compared.  The comparative analysis, show features clearly distinguishing these two types of QAP instances. Apart from a clear confirmation that the search difficulty increases with the problem dimension, the results provide new confirming evidence explaining  why the real-like instances are easier to solve exactly, while the uniform instances are easier to solve approximately using stochastic local search.

A comparison of  the LON of QAP against those of the previously studied  $NK$ landscapes~\cite{gecco08,alife08,pre09}, suggests that the distributions of basin sizes, including the global optimum basin, are similar for comparable instances of QAP and $NK$. The main difference lies in the connectivity: whereas the LON is nearly a complete graph for QAP, this is not the case for  $NK$ landscapes. This could  be due to the larger neighborhood size in the permutation space, as compared with the binary space. With the current move operations, the probability of exploring another basin from a solution by a random move is higher for comparable QAP and $NK$ instances. This suggests that the most efficient local searcher (based on those moves), for each problem should be different: the tradeoff between exploration and exploitation should not be the same. The aforementioned comparison could also permit to better explain the parallel between some flow matrix characteristics of QAP and the epistasis value of $NK$. In particular, the influence of {\em flow dominance}, a metric often used to characterize QAP instances, has not been directly addressed in this paper, and it surely deserves more attention and space.
There may be also a connection between the present work and the concept of \textit{elementary landscapes}, as
QAP spaces or parts thereof might be spectrally decomposable in such way~\cite{stadler-02}. 
%\enlargethispage{-0.45in}
\enlargethispage{0.16cm}

There are several directions for future work. Our current definition of transition probabilities, although very informative, may produce highly connected networks, which are not easy to study. Therefore, we are currently considering alternative definitions based on threshold values for the connectivity. A high variability (across instances) on some metrics was observed, especially for the real-like class. Therefore, further analysis may need to be focused on particular instances, instead of on a statistical aggregation of a set of instances of (arguably) the same class. Moreover, our current methodology is only applicable to small problem instances. Good sampling techniques are required in order to extend the applicability of the model. The consideration of a permutation space in this article, opens up the possibility of analyzing other permutation based problems such as the traveling salesman and the permutation flow shop problems. In addition, it would be useful to
compare results based on LON representation with those arising from theoretical analyses such as~\cite{Chen2010926}.

Finally, although the local optima network model is still under development, we argue that it offers an alternative view of combinatorial fitness landscapes, which can potentially contribute to both  our understanding of problem difficulty, and the design of effective heuristic search algorithms, including evolutionary algorithms.

\section*{Acknowledgment}

Fabio Daolio and Marco Tomassini
gratefully acknowledge the Swiss National Science Foundation for
financial support under grant number 200021-124578.

\end{document}